\documentclass[conference]{IEEEtran}
\IEEEoverridecommandlockouts
\usepackage{cite}
\usepackage{amsmath,amssymb,amsfonts}
\usepackage{algorithmic}
\usepackage{graphicx}
\usepackage{textcomp}
\usepackage{xcolor}
\def\BibTeX{{\rm B\kern-.05em{\sc i\kern-.025em b}\kern-.08em
    T\kern-.1667em\lower.7ex\hbox{E}\kern-.125emX}}
\begin{document}

\title{Integrating Deep RL and Bayesian Inference for ObjectNav in Mobile Robotics\\
}

\author{\IEEEauthorblockN{João Castelo-Branco}
 \IEEEauthorblockA{\textit{Instituto Superior Técnico}\\
 Lisbon, Portugal \\
 joao.castelo.branco@tecnico.ulisboa.pt}
 \and
 \IEEEauthorblockN{José Santos-Victor}
 \IEEEauthorblockA{\textit{Institute For Systems and Robotics}\\
 Lisbon, Portugal \\
 jasv@isr.ist.utl.pt}
 \and
 \IEEEauthorblockN{Alexandre Bernardino}
\IEEEauthorblockA{
 \textit{Institute For Systems and Robotics}\\
 Lisbon, Portugal \\
 alex@isr.ist.utl.pt}
 
 %\and
% \and
% \and
}

\maketitle

\begin{abstract}
Autonomous object search is challenging for mobile robots operating in indoor environments due to partial observability, perceptual uncertainty, and the need to trade off exploration and navigation efficiency. Classical probabilistic approaches explicitly represent uncertainty but typically rely on handcrafted action-selection heuristics, while deep reinforcement learning enables adaptive policies but often suffers from slow convergence and limited interpretability. This paper proposes a hybrid object-search framework that integrates Bayesian inference with deep reinforcement learning. The method maintains a spatial belief map over target locations, updated online through Bayesian inference from calibrated object detections, and trains a reinforcement learning policy to select navigation actions directly from this probabilistic representation. The approach is evaluated in realistic indoor simulation using Habitat 3.0 and compared against developed baseline strategies.
%, including a purely probabilistic belief-driven method. 
Across two indoor environments, the proposed method improves success rate while reducing search effort. 
%In the larger environment, it achieves a $+5$ percentage point increase in success rate and reduces the average number of executed actions and traveled distance by $23\%$ and $18\%$, respectively, relative to the probabilistic baseline. 
Overall, the results support the value of combining Bayesian belief estimation with learned action selection to achieve more efficient and reliable object-search behavior under partial observability.
\end{abstract}


\section{Introduction}

\begin{figure}[t]
    \centering
    \includegraphics[width=\columnwidth]{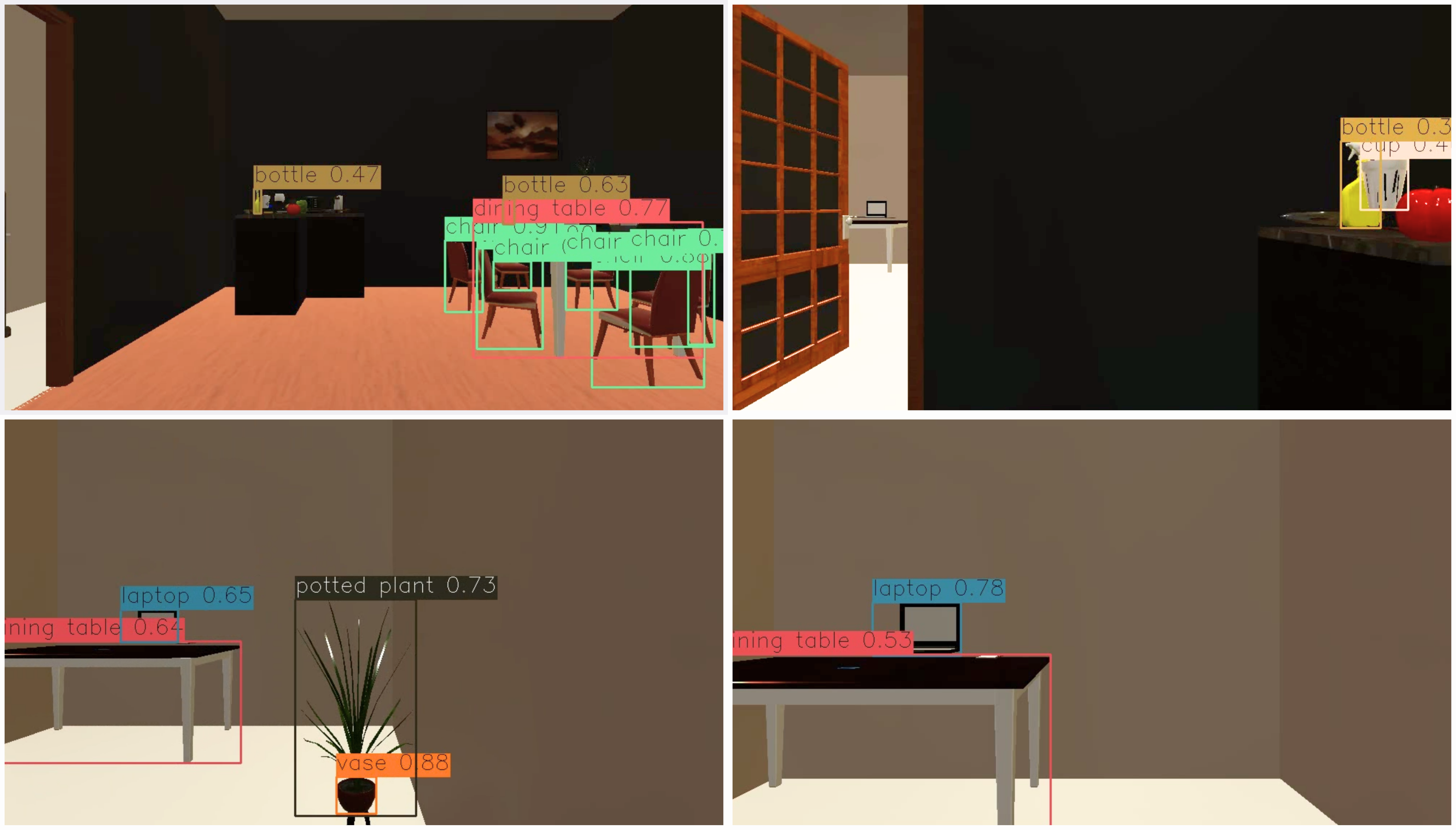}
    \caption{Example object-search episode in a domestic indoor environment. 
    The robot is tasked with locating a \emph{laptop} and declaring success when the detection confidence exceeds $75\%$. 
    (top-left) Initial exploration in the living room, where the target is not detected. 
    (top-right) The robot explores an adjacent room through the doorway; although the laptop is visible from an oracle perspective, it is not yet identified by the detector. 
    (bottom-left) After entering the room, the robot detects the laptop but with insufficient confidence to terminate the task. 
    (bottom-right) The robot actively moves closer to reduce perceptual uncertainty, achieving a confident detection and successfully completing the search.}

    \label{fig:search_example}
\end{figure}

Object search is a fundamental capability for autonomous mobile robots operating in complex, unstructured environments
%, requiring the localization of target objects under incomplete and uncertain sensory information 
~\cite{rasouli2020attention}. The problem is challenging due to partial observability, high-dimensional state spaces, and perceptual uncertainty arising from limited sensor coverage, occlusions, and noisy detections~\cite{problem-definition}. Fig.~\ref{fig:search_example} illustrates a representative object-search episode under partial observability, highlighting the progression from exploration to confident target detection.

Existing research on robotic object search has largely followed two complementary directions. Classical probabilistic approaches address uncertainty by maintaining an explicit belief distribution over potential target locations and updating it as new perceptual evidence becomes available~\cite{thrun2002probabilistic}. This enables interpretable uncertainty modeling and principled decision-making under partial observability. However, action selection is typically driven by handcrafted heuristics or predefined utility functions, which can limit scalability and adaptability across diverse environments and sensing conditions.

In contrast, deep reinforcement learning (DRL) methods learn policies directly from experience, enabling agents to acquire effective exploration behaviors without manual feature engineering~\cite{ye_active_2018}. Despite strong empirical results, standard end-to-end DRL approaches often suffer from slow convergence and inefficient exploration, and they commonly lack explicit uncertainty reasoning and interpretability~\cite{chaplot2020object}.

To address these limitations, this paper proposes a hybrid framework that integrates Bayesian inference with DRL for efficient object search under partial observability. The method maintains a spatial belief map representing the probabilistic distribution of target locations, updated online through Bayesian inference using calibrated object detections. A DRL policy is trained to select actions that balance exploration and exploitation based on the current belief state. By combining explicit uncertainty modeling with data-driven policy learning, the framework provides a unified decision-making architecture that couples interpretability with adaptability. 
%To the best of our knowledge, this integration has not been previously reported in object search tasks.

The proposed framework is evaluated in the Habitat~3.0 simulator~\cite{puig_habitat_2023} across multiple indoor environments 
%and compared against baselines. A cylindrical 
using a mobile robot 
%model 
equipped with an RGB-D sensor and geometric mapping, navigation and localization skills.
%is used for the experiments.

\section{Related Work}
\label{sec:related_work}

Robotic object search is closely related to active perception, where an agent must decide how to move in order to acquire informative observations while minimizing search cost~\cite{rasouli2020attention}. Earlier work in this area relied on geometric reasoning, symbolic representations, and handcrafted perception pipelines~\cite{shubina_visual_2010}. With the rise of deep neural networks for detection and semantic labeling, object search has increasingly shifted toward data-driven perception and learning-based decision making. As a result, existing approaches can be broadly grouped into two methodological paradigms: \textbf{probabilistic belief-based methods} and \textbf{learning-based methods}. These paradigms have evolved largely in parallel and address complementary aspects of the problem: probabilistic approaches emphasize explicit uncertainty modeling and structured decision making, while reinforcement learning approaches focus on learning adaptive policies through interaction. 

\textbf{Probabilistic approaches} represent the target location as a belief distribution that is iteratively updated as new observations are acquired, typically through Bayesian filtering~\cite{thrun2002probabilistic}. These methods provide interpretable uncertainty representations and support decision-making under partial observability by selecting actions that maximize expected utility or information gain. Several works incorporate semantic or structural priors to improve scalability in indoor environments, for example by reasoning over spatial relations, hierarchical maps, or metric--topological abstractions~\cite{2013-Uncertain-Semantics,veiga_efficient_2016,wang_efficient_2018,zhang_building_2022}. Despite their principled uncertainty handling, action selection is often driven by subjective heuristics or manually designed utility functions, which may require environment-specific tuning and can limit adaptability in diverse settings.

\textbf{Learning-based approaches}, primarily based on DRL, learn search and navigation policies directly from interaction with the environment and guided by an objective reward signal (e.g., upon task success), enabling agents to acquire effective exploration behaviors without manual feature engineering~\cite{zhu_target-driven_2017,ye_active_2018}. Many methods improve performance by shaping rewards, leveraging semantic mapping, or combining learned policies with classical planning components to guide exploration~\cite{chaplot2020object}. Additional strategies incorporate auxiliary learning objectives to enhance sample efficiency and robustness~\cite{ye2021auxiliary}. While these approaches can achieve strong empirical results, they typically lack explicit 
%probabilistic reasoning about 
representation of
perceptual uncertainty, reducing interpretability and potentially leading to inefficient exploration under noisy detections. Moreover, training requires substantial interaction data and computational resources.

In summary, probabilistic models offer explicit uncertainty representation but commonly rely on handcrafted decision rules, whereas DRL-based methods learn adaptive behaviors but largely omit principled belief modeling. This work addresses this gap by integrating Bayesian belief updates with DRL-based action selection for uncertainty-aware object search. In addition, we develop baseline strategies, including a purely probabilistic belief-driven method with a handcrafted utility-maximization policy, enabling an objective evaluation of the benefits introduced by learning-based control.

\section{Proposed Approach}
\label{sec:method}

\begin{figure}[t]
    \centering
    % Replace with thesis figure or paper-specific diagram
    % 
    \includegraphics[width=0.7\columnwidth]{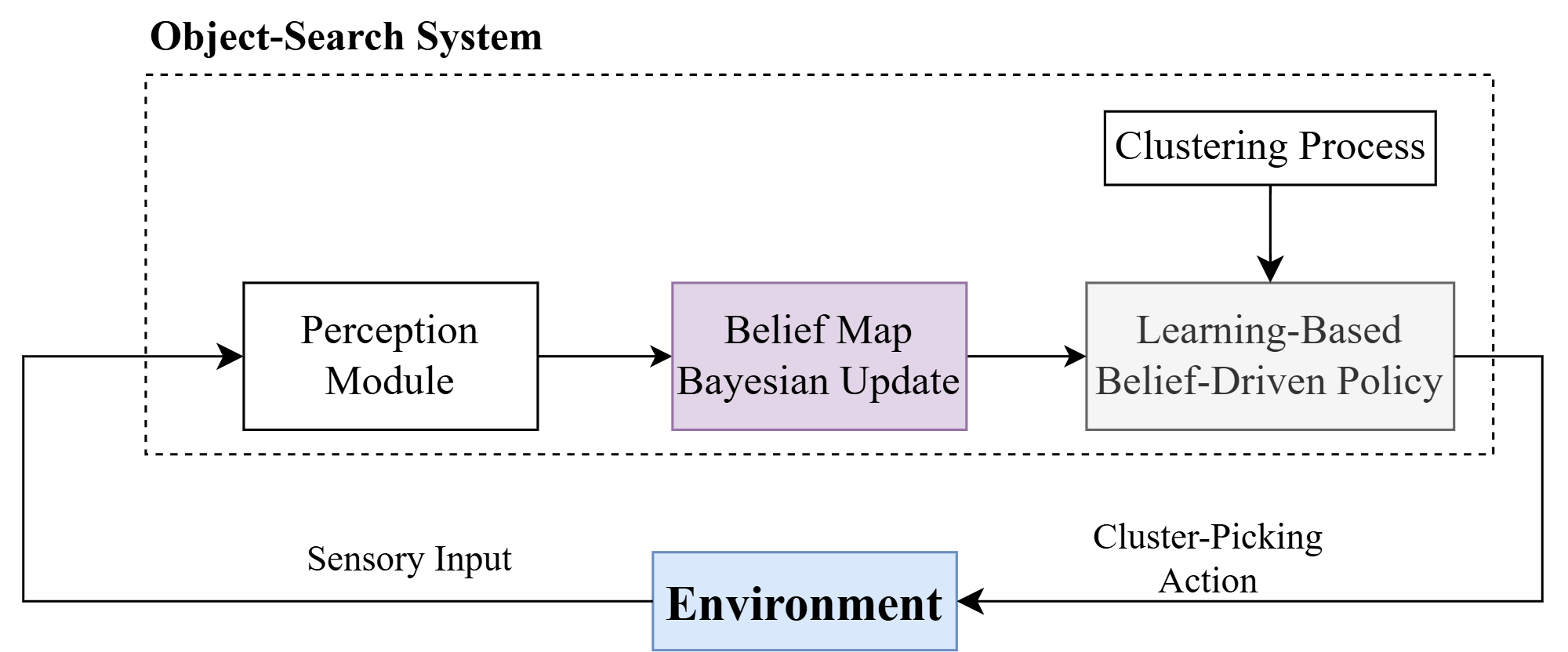}
    \caption{Overview of the proposed hybrid object-search framework. The robot converts RGB-D observations into spatial evidence, updates a Bayesian belief map over target locations, and selects actions using a DRL policy conditioned on the belief representation. A clustering subsystem partitions free space into candidate regions to structure exploration and navigation.}
    \label{fig:system_overview}
\end{figure}

Fig.~\ref{fig:system_overview} presents an overview of the proposed hybrid framework for active object search under partial observability. The key idea is to combine explicit probabilistic belief estimation with learning-based decision making: the robot maintains a spatial belief map over the target location, updated online from perceptual evidence, and uses a DRL policy to select navigation actions conditioned on this belief representation. 
%It is assumed that the robot has navigation and self-localization capabilities. 
The robot is provided with a geometric map of the environment \textit{a priori} in the form of a 2D occupancy grid map, where objects can only occupy occupied cells, i.e., non-navigable space. The semantic content of these cells must be inferred.

At each time step, the robot acquires an RGB-D observation and applies an object detection pipeline to obtain candidate detections for the target category. These detections are transformed into spatial evidence by projecting them into the map, producing an observation signal that is consistent with the robot pose and the scene geometry. The belief map is then updated through Bayesian inference, integrating the new evidence with the prior belief to maintain a probabilistic estimate of where the target is likely to be located.

To support efficient exploration in large environments, the free space of the known binary occupancy grid is also partitioned into a set of spatial clusters. This clustering subsystem provides a compact representation of candidate viewpoints and structures the navigation space, allowing the agent to reason and act over meaningful regions rather than individual grid cells. The belief representation and action selection are therefore defined with respect to this clustered map structure.

The resulting belief state serves as a compact and uncertainty-aware representation of the search problem. Rather than relying on a handcrafted utility function to decide where to move next, a DRL policy is trained to map the current belief representation to an action in a discrete motion space. This allows the agent to learn search behaviors that balance exploration and exploitation while leveraging the structure provided by Bayesian uncertainty modeling.

The following subsections describe the main building blocks of the framework: (i) the perception and observation model, (ii) the belief representation and Bayesian update rule, (iii) the clustering and navigation abstraction, and (iv) the DRL-based policy that selects actions from the belief state.

% -------------------- FIGURE PLACEHOLDER --------------------
% ------------------------------------------------------------

\subsection{Perception and Observation Model}
\label{sec:perception}

The perception module 
%(see Fig.~\ref{fig:perception_pipeline}) 
provides a shared sensory front-end for all methods evaluated in this work. Its role is to transform raw RGB-D observations into metrically grounded and uncertainty-aware evidence that can be integrated into the Bayesian belief update. At each time step $t$, the robot receives $O_t = \left(I^{\text{RGB}}_t, I^{D}_t\right)$, where $I^{\text{RGB}}_t$ is the RGB image and $I^{D}_t$ is the aligned depth map. The robot is assumed to have access to the environment map $M$ (i.e., free space and obstacles), but the semantic contents of occupied cells are unknown and must be inferred from perception over time.

Object detections are obtained from $I^{\text{RGB}}_t$ using the YOLO~\cite{yolo-paper} object detector (v11). Each detection produces a bounding box and a confidence distribution over object classes. To ensure these confidences can be interpreted probabilistically, the standard multi-label sigmoid activation in the classification head is replaced by a softmax layer, enforcing mutually exclusive class probabilities. In addition, the output probabilities are calibrated by tuning the softmax temperature parameter $T$ on the MS~COCO dataset~\cite{coco-paper}, in order to improve consistency between predicted confidence and empirical accuracy. This step reduces overconfidence and provides more reliable evidence for belief updates.

To integrate detections into spatial reasoning, each detection is projected from image coordinates into the environment map using the depth observation and the known camera--robot geometry. A depth estimate at the detection center is used to compute a 3D point in the camera frame, which is then transformed into world coordinates using the robot pose $P_t$. The resulting $(x,y)$ location is discretized into the 2D occupancy grid $M$, yielding a map-level observation associated with a specific spatial cell. When the projected location falls in free space, it is reassigned to the nearest occupied cell, reflecting the assumption that objects can only be located in non-traversable regions of the map.

The output of this module is a set of spatially grounded, calibrated probabilistic observations that serve as  inputs to the Bayesian belief update described in the next subsection.

% -------------------- FIGURE PLACEHOLDER --------------------
%\begin{figure}[t]
%    \centering
    % Replace with thesis figure (perception pipeline / projection diagram)
    % 
%    \includegraphics[width=\columnwidth]{Figures/perception_module.png}
%    \caption{Perception pipeline used to generate spatial evidence from RGB-D observations. YOLOv11 detections are calibrated and projected into the occupancy grid to support Bayesian belief updates.}
%    \label{fig:perception_pipeline}
%\end{figure}
% ------------------------------------------------------------

\subsection{Belief Representation and Bayesian Update}
\label{sec:belief_update}

To explicitly model uncertainty about the target location under partial observability, the proposed framework maintains a probabilistic belief map over the environment. The robot is assumed to know the occupancy grid $M$ (free space vs.\ obstacles), but the semantic contents of occupied cells are unknown and must be inferred from observations. We therefore associate each occupied cell $(i,j)$ with a Dirichlet distribution over $K$ object classes plus a background class:
\begin{equation}
    \boldsymbol{\beta}_{i,j}
    =
    \left[\beta^{(1)}_{i,j},\ldots,\beta^{(K)}_{i,j},\beta^{(bg)}_{i,j}\right]^{\top}
    \in \mathbb{R}^{K+1}_{>0}.
\end{equation}
The full belief map is denoted by $B=\{\boldsymbol{\beta}_{i,j}\}$, and the posterior mean categorical distribution at each cell is
\begin{equation}
\hat{\pi}^{(k)}_{i,j}
=
\beta^{(k)}_{i,j} \Big / 
\left( \sum_{m=1}^{K+1}\beta^{(m)}_{i,j}
\right),
\quad
k \in \{1,\ldots,K,bg\}.
\end{equation}

At initialization, an uniform prior is used, i.e., $\boldsymbol{\beta}_{i,j}=\mathbf{1}$.

\textit{a) Observation evidence.}
At time $t$, the perception module (Sec.~\ref{sec:perception}) produces calibrated detections mapped to grid cells. Each detection yields a class-probability vector
$p \in \Delta^{K-1}$ over the $K$ object classes, which is converted into a cell-level observation vector
$o \in \Delta^{K}$ by allocating a fixed probability mass to background and rescaling the remaining mass across object classes:
\begin{equation}
    o_k = p_k \, K/(K+1),
    \qquad
    o_{bg} = 1/(K+1).
    \label{eq:obs_from_det}
\end{equation}
This prevents overly confident updates from single detections and ensures that all evidence remains compatible with the Dirichlet belief representation.

In addition to positive detections, the absence of a detection in visible occupied cells provides weak negative evidence favoring the background hypothesis. Let $G^{bg}_t$ denote the set of occupied cells that are visible at time $t$ but have no mapped detections. For each $(i,j)\in G^{bg}_t$, we construct an observation vector $o\in\Delta^K$ whose background mass decays with the 2D distance $\rho$ between the robot and the cell center,
$
    o_{bg} = (1-\eta)/(1+\lambda \rho),
$
where $\eta$ is the detector false-negative rate and $\lambda>0$ controls distance decay. The remaining probability mass is uniformly distributed across object classes, 
$
    o_k = (1-o_{bg})/K,\quad k=1,\ldots,K.
$

\begin{figure}[t]
    \centering
    \includegraphics[width=\columnwidth]{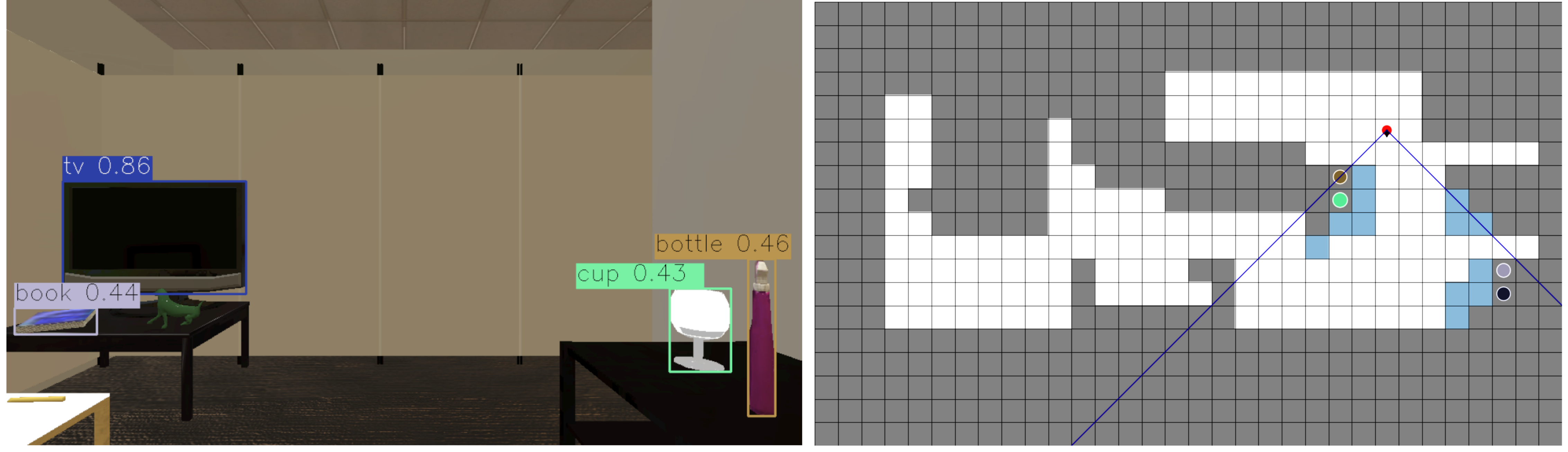}
    \caption{Example of map-level evidence generation at a single time step. \textbf{Left:} RGB frame with detections; only the most likely class label per bounding box is shown, while the full categorical output vector $p$ is used to compute observation evidence. \textbf{Right:} detections projected onto the occupancy grid; each projected detection is represented by a colored circle matching the corresponding bounding box. Blue cells denote occupied map cells within the robot field of view (blue rays) that have no mapped detections at that time step, providing background evidence for the Bayesian belief update.}
    \label{fig:belief_evidence_mapping}
\end{figure}

Fig.~\ref{fig:belief_evidence_mapping} illustrates how positive and background evidence are mapped to the occupancy grid at a single time step.

% -------------------- FIGURE: EVIDENCE MAPPING --------------------
% ---------------------------------------------------------------

\textit{b) Bayesian fusion.}
Beliefs are updated online by fusing the prior Dirichlet parameters with the new observation evidence. While a standard conjugate update would directly accumulate pseudo-counts, this can lead to overly confident posteriors under noisy detections. Instead, we adopt the conservative fusion rule proposed by Kaplan \textit{et al.}~\cite{kaplan}. Given a prior parameter vector $\boldsymbol{\beta}$ for a cell and an observation vector $o$, the updated parameters are
\begin{equation}
\beta_k^{+}
=
\beta_k
\,
\frac{\sum_{j=1}^{K+1}\beta_j o_j + o_k}
     {\sum_{j=1}^{K+1}\beta_j o_j + \min_i o_i},
\quad k = 1,\ldots,K+1 .
\label{eq:kaplan_update}
\end{equation}

This update is applied to the corresponding grid cell for each mapped observation. The resulting belief map $B$ provides a compact and uncertainty-aware representation of the search problem, and constitutes the main state input to the action-selection module described next.

\subsection{Belief-Driven Policy with Clustering Abstraction}
\label{sec:policy_learning}

The belief map $B_t$ (Sec.~\ref{sec:belief_update}) provides an explicit representation of the agent's hypothesis about the target location and its uncertainty at time $t$. We use this belief to define a compact decision process over clustered navigation goals and learn a policy with deep Q-learning.

\textit{a) Belief-state representation.}
The DQN input is the tensor
$    \mathcal{T}_t =
    \left[
        \hat{\pi}^{(\kappa)}_t,\ 
        H_t,\ 
        M,\ 
        \delta(X_t,Y_t)
    \right] \in \mathbb{R}^{C\times H\times W}
$
($C=4$) where $\hat{\pi}^{(\kappa)}_t \in [0,1]^{H\times W}$ is the posterior mean probability of the target class $\kappa$ at each cell (Sec.~\ref{sec:belief_update}). Let $H_t \in [0,1]^{H\times W}$ be the normalized categorical entropy map computed from the corresponding posterior mean distribution, capturing local uncertainty. The occupancy grid $M\in\{0,1\}^{H\times W}$ encodes the free/occupied structure of the environment, and $\delta(X_t,Y_t)$ is a one-hot map indicating the current robot position.

\begin{figure}[t]
    \centering
    \includegraphics[width=0.55\columnwidth]{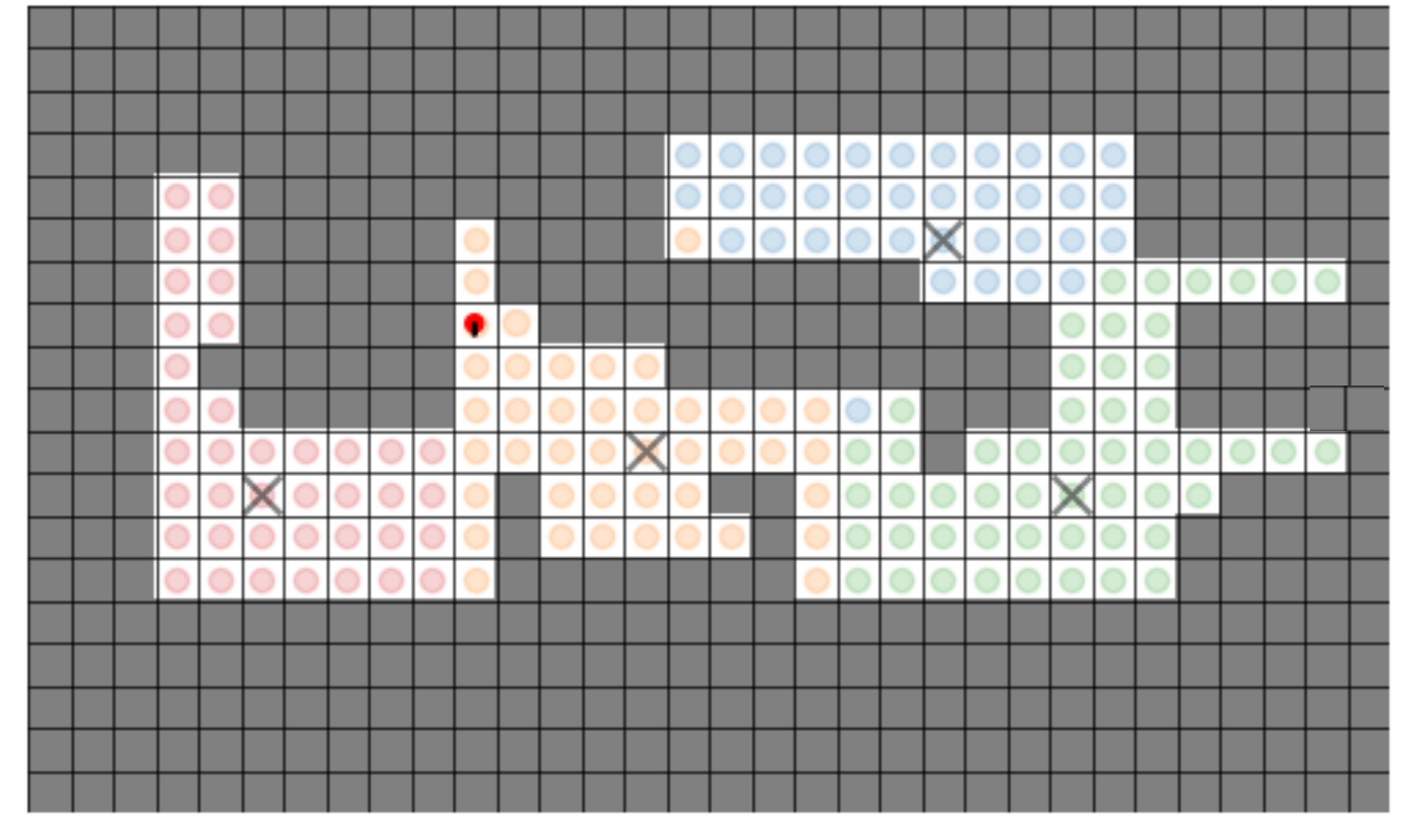}
    \caption{Clustering-based abstraction of the navigation space. Free cells are partitioned into spatial clusters (colored regions), each represented by a centroid (marked with $\times$) used as a candidate viewpoint for exploration.}
    \label{fig:map_clustering}
\end{figure}

\textit{b) Clustering and progressive refinement.}
Let $F$ denote the set of free cells in $M$. At refinement level $\ell$, we partition $F$ into $k_\ell$ spatial clusters, each represented by a centroid used as a candidate viewpoint (Fig.~\ref{fig:map_clustering}). At each decision step, the agent selects one centroid to visit and navigates to it; this repeats until the target is found or the time horizon $T$ is reached. If all centroids at level $\ell$ are visited without success, we refine the partition by doubling the number of clusters $k_\ell = \min(2^\ell k_0,\ |F|)$.

% -------------------- FIGURE --------------------
% -----------------------------------------------

\textit{c) DQN over cluster centers.}
We train a Deep Q-Network (DQN) to approximate the optimal action-value function over navigation goals. Here, the decision variable is a \emph{goal} $g$ (rather than a primitive action), defined as a grid cell corresponding to a target cluster centroid. Formally,
\begin{equation}
    Q^\ast(\mathcal{T},g) = \max_{\pi} \mathbb{E}_{\pi}\!\left[\sum_{t=0}^{T} \gamma^t r_t \,\middle|\, \mathcal{T}_0=\mathcal{T},\ g_0=g \right],
\end{equation}
with neural approximation $Q_\theta(\mathcal{T},g)$. The network outputs a dense Q-map over free cells, $Q(\mathcal{T}_t;\theta)\in \mathbb{R}^{|F|}$,
interpreted as the expected return of selecting each free cell as the next goal. To enforce the clustering abstraction at level $\ell$, we restrict goals to the set of admissible centroids $\mathcal{U}_\ell$ using a binary mask $M_\ell \in \{0,1\}^{|F|}$:
\begin{equation}
    \tilde{Q}_{i,j} =
    \begin{cases}
        Q_{i,j}(\mathcal{T}_t;\theta), & \text{if } M_\ell[(i,j)]=1,\\
        -\infty, & \text{otherwise.}
    \end{cases}
    \label{eq:q_mask}
\end{equation}
The next goal $g_t$ is selected using an $\varepsilon$-greedy policy over
$\mathcal{U}_\ell=\{(i,j)\in F \mid M_\ell[(i,j)]=1\}$, choosing a uniformly
random element of $\mathcal{U}_\ell$ with probability $\varepsilon$ and
$\arg\max_{(i,j)\in\mathcal{U}_\ell}\tilde{Q}_{i,j}$ with probability $1-\varepsilon$.

\textit{d) Execution and reward.}
Executing a high-level action corresponds to planning and traversing a shortest path to $g_t$ using motion primitives, while continuously collecting RGB--D observations and updating $B_t$ online (Sec.~\ref{sec:belief_update}). Rewards are assigned at the goal-selection level and aggregate the cost of executed primitives:
$
    r_t = n_{\text{prim}}\cdot r_{\text{step}} + \mathbf{1}_{\text{succ}}\cdot r_{\text{succ}},
$
where $n_{\text{prim}}$ is the number of primitives executed until the next decision point (or until success), and $\mathbf{1}_{\text{succ}}$ indicates whether the target was found during this segment. We set $r_{step}=-0.01$ and $r_{succ}=1$.

\section{Experimental Setup}
\label{sec:experiments}

\subsection{Simulation Environment and Evaluation Metrics}
\label{sec:sim_and_task}
All experiments are conducted in two realistic indoor scenes simulated in Habitat~3.0: a smaller office workspace ($9.76 \times 5.95 \times 2.71$~m) and a larger two-bedroom apartment ($13.78 \times 13.60 \times 3.21$~m). The agent is equipped with RGB--D sensing (resolution $1024 \times 576$) and navigates using a discrete motion model with the primitives \texttt{move\_forward}, \texttt{turn\_left}, and \texttt{turn\_right}. The camera horizontal field-of-view is $90^\circ$. The occupancy grid $M$ is assumed to be known at a resolution of $0.30$~m, while the semantic contents of occupied cells are unknown and must be inferred online through observations (Sec.~\ref{sec:belief_update}). The robot is modeled with collision radius $0.15$~m and height $1.0$~m.

Each episode involves searching for a single instance of a target object class within a fixed environment (only one such instance is present per environment). The robot starts from a predefined initial pose and iteratively selects navigation goals (Sec.~\ref{sec:policy_learning}), updating its belief map as new observations become available. An episode terminates when the target is detected above a confidence threshold $\tau=0.8$ or when a maximum planning horizon $T$ is reached. 

Performance is evaluated in terms of both \emph{effectiveness} and \emph{efficiency}. Effectiveness is measured by the \emph{success rate} (SR), defined as the fraction of episodes in which the target is found within the time horizon. To quantify search efficiency, we report the average number of executed primitive actions and the traveled distance per episode. These metrics capture the ability of a method to localize the target reliably while minimizing search effort.

\subsection{Baselines and Comparative Methods}
\label{sec:baselines}
We compare the proposed Bayesian Belief-Driven Policy Search (BBDPS) against a set of baselines designed to isolate the contribution of each component. All methods operate under the same assumptions and experimental conditions, including the same known occupancy grid $M$, the same perception module, and the same episode termination criteria.

\textit{a) Random Walk Search (RWS):}
A baseline that selects motion primitives randomly, yielding an uninformed exploration strategy with no explicit use of the belief representation.

\textit{b) Progressive Cluster Sweep Search (PCSS).}
PCSS uses the same clustering and progressive refinement mechanism described in Sec.~\ref{sec:policy_learning}, but follows a deterministic sweeping strategy: cluster centroids are visited sequentially in a greedy distance-based order to encourage coarse-to-fine coverage of the environment. This baseline represents structured exploration without probabilistic reasoning or learning.

\textit{c) Bayesian Belief--Utility Maximization Search (BBUMS).}
BBUMS represents a purely probabilistic, belief-driven approach. It maintains the same belief map $B_t$ and selects the next navigation goal by maximizing a handcrafted utility function defined over the current belief state, trading off target likelihood, uncertainty, and motion cost. The next cluster center $c^*$ is selected as:
\begin{equation}
c^* = \arg\max_{m} \quad w_H \bar{H}_m + w_d \left(1 - \bar{d}_m\right) + w_p \bar{p}_m,
\end{equation}
\noindent where $\bar{H}_m$ is the normalized average entropy over cluster $c_m$, $\bar{p}_m$ is the normalized maximum posterior probability of the target class $\kappa$ across all cells in cluster $c_m$, and $\bar{d}_m$ is the normalized motion cost to the cluster center. We set $w_p = 0.1$, $w_d = 0.5$, and $w_H = 0.4$. This baseline serves as a proxy for classical Bayesian object-search strategies that rely on explicit uncertainty modeling but fixed decision rules.

\textit{d) Bayesian Belief-Driven Policy Search (BBDPS).}
Our method replaces handcrafted action selection with a learned policy: a DQN is trained to select cluster-centroid goals directly from the belief-state tensor $\mathcal{T}_t$ 
%(Sec.~\ref{sec:policy_learning}), enabling 
for adaptive decision making conditioned on accumulated probabilistic evidence.

\subsection{Policy Training Protocol}
\label{sec:training_details}
\begin{figure}[t]
    \centering
    \includegraphics[width=0.5\columnwidth]{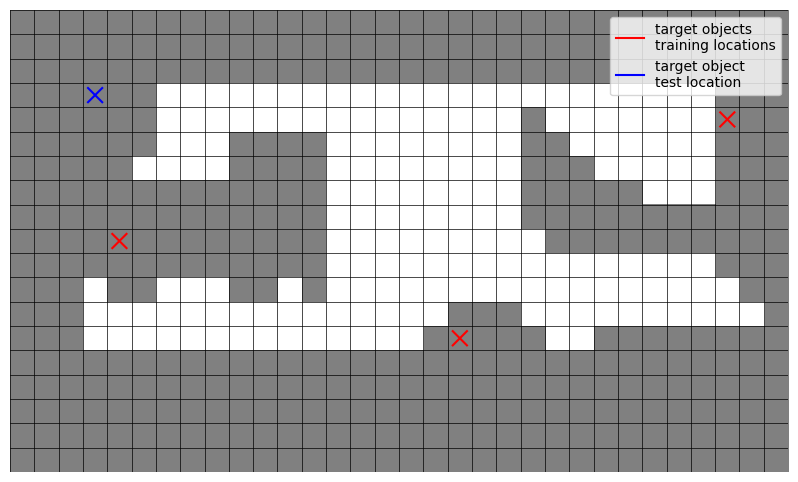}
    \caption{Training and evaluation targets in Env.~1. The policy is trained on a \textit{potted plant}, \textit{laptop}, and \textit{teddy bear}, and evaluated on a held-out \textit{tv}.}

    \label{fig:env1_targets}
\end{figure}

\begin{figure}[t]
    \centering
    \includegraphics[width=0.55\columnwidth]{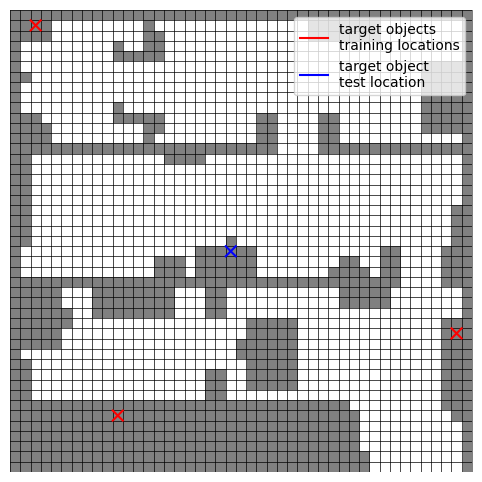}
    \caption{Training and evaluation targets in Env.~2. The policy is trained on a \textit{potted plant}, \textit{couch}, and \textit{toilet}, and evaluated on a held-out \textit{laptop}.}

    \label{fig:env2_targets}
\end{figure}

Since the policy input $\mathcal{T}_t$ is defined over the environment grid (Sec.~\ref{sec:policy_learning}), independent DQNs are trained for each environment.
For each environment, training episodes are generated by sampling target placements from a predefined set. At the start of each episode, one of three possible target instances from different object classes is selected, each placed at a fixed location, and the robot explores the environment until either the target is found or the planning horizon is reached. Evaluation follows the same protocol but uses a different single held-out target instance per environment (Figs.~\ref{fig:env1_targets}--\ref{fig:env2_targets}). 

Training hyperparameters are fixed across environments, including discount factor $\gamma=0.99$, learning rate $10^{-3}$, batch size $64$, replay buffer capacity $50{,}000$, target network update frequency $2000$ steps, and $5000$ training episodes per environment. The exploration parameter $\varepsilon$ is annealed during training from $1.0$ to $0.05$.

% -------------------- FIGURES --------------------

% -----------------------------------------------

\section{Results and Discussion}
\label{sec:results}

For each algorithm--environment pair, we execute $100$ test episodes starting from distinct initial robot poses. The set of initial poses is identical across all methods, ensuring strictly comparable starting conditions.

\subsection{Success Rate and Search Efficiency}
\label{sec:quant_results}

Table~\ref{tab:sr_both_envs} reports the success rate (SR) in both environments for a time horizon of $T=0.75N$, where $N$ denotes the number of free grid cells in the environment. Among all methods, the belief-driven approaches (BBUMS and BBDPS) attain the highest SR, with BBDPS achieving perfect success in Env.~1 and near-perfect success in Env.~2.

\begin{table}[t]
\centering
\caption{Success rates across environments.}
\label{tab:sr_both_envs}
\begin{tabular}{lcccc}
\hline
\textbf{Env.} & \textbf{RWS} & \textbf{PCSS} & \textbf{BBUMS} & \textbf{BBDPS} \\
\hline
Env.~1 & 0.52 & 0.85 & 0.87 & \textbf{1.00} \\
Env.~2 & 0.43 & 0.89 & 0.94 & \textbf{0.99} \\
\hline
\end{tabular}
\end{table}

To compare search effort under comparable outcomes, we additionally report efficiency metrics computed only over \emph{joint-success} episodes, i.e., the subset of test runs where \emph{all} methods successfully found the target. This avoids biasing action and distance averages for weaker baselines, which may succeed only on easier episodes while failing on more difficult ones. In Env.~1 and Env.~2, the number of joint-success episodes is $41$ and $40$ (out of $100$), respectively. Table~\ref{tab:eff_env1} reports the mean and standard error of the number of executed actions and traveled distance for each environment.

\begin{table}[t]
\centering
\caption{Efficiency (mean $\pm$ SE) over joint-success episodes.}
\label{tab:eff_env1}
\begin{tabular}{lcc}
\hline
\textbf{Method} & \textbf{Actions (Env.~1)} $\downarrow$ & \textbf{Dist. (m) (Env.~1)} $\downarrow$ \\
\hline
RWS   & $37.56 \pm 5.55$ & $3.05 \pm 0.47$ \\
PCSS  & $\mathbf{12.17 \pm 1.18}$  & $\mathbf{1.91 \pm 0.22}$ \\
BBUMS & $13.00 \pm 1.45$  & $2.07 \pm 0.28$ \\
BBDPS & $20.66 \pm 4.23$ & $4.05 \pm 0.89$ \\
\hline
\end{tabular}

\vspace{0.6em}
\label{tab:eff_env2}
\begin{tabular}{lcc}
\hline
\textbf{Method} & \textbf{Actions (Env.~2)} $\downarrow$ & \textbf{Dist. (m) (Env.~2)} $\downarrow$ \\
\hline
RWS   & $244.33 \pm 42.34$ & $\mathbf{21.46 \pm 3.56}$ \\
PCSS  & $235.10 \pm 21.92$ & $49.99 \pm 4.55$ \\
BBUMS & $140.85 \pm 13.05$  & $31.45 \pm 3.07$ \\
BBDPS & $\mathbf{109.10 \pm 11.59}$ & $25.67 \pm 2.75$ \\
\hline
\end{tabular}
\end{table}

\subsection{Discussion}
\label{sec:discussion}

\textit{a) Belief-driven methods improve reliability:}
Across both environments, BBUMS and BBDPS achieve the highest success rates (Table~\ref{tab:sr_both_envs}), supporting the importance of explicitly tracking probabilistic hypotheses over target locations under partial observability. Compared to PCSS, belief-driven approaches can re-prioritize navigation goals online based on accumulated evidence rather than geometric coverage.

\textit{b) Learning yields the largest efficiency gains in the larger environment.}
In Env.~2, BBDPS achieves the lowest average number of actions among all methods on joint-success episodes ($109.10$ vs.\ $140.85$ for BBUMS), corresponding to a reduction of approximately $23\%$. The traveled distance is also reduced relative to BBUMS ($25.67$ vs.\ $31.45$\,m, about $18\%$). These gains indicate that the learned goal-selection policy improves long-horizon decision making by exploiting the belief structure more effectively than a fixed handcrafted utility, particularly in larger maps where exploration order has a stronger impact on search cost.

\textit{c) Small environments reduce the advantage of learned goal ordering.}
In Env.~1, BBDPS attains perfect success, but its efficiency on joint-success episodes is worse than PCSS and BBUMS (Table~\ref{tab:eff_env1}). This suggests that in compact scenes, deterministic coverage or utility-based goal selection is already effective, leaving less room for improvement through learning. The larger standard error in BBDPS distance further indicates that the learned policy can occasionally commit to longer trajectories before converging to the correct region.

\textit{d) Structured exploration alone is insufficient.}
PCSS substantially improves over random walk, confirming that clustering and coarse-to-fine coverage are strong inductive biases for scalable search. However, PCSS lacks belief awareness and may traverse large portions of the map even when evidence already suggests more informative regions, which is reflected in its high traveled distance in Env.~2 (Table~\ref{tab:eff_env1}).

\textit{e) Failure modes.}
Remaining failures are typically perception-limited: if the target remains occluded for long periods or detections are weak/ambiguous, the belief can remain diffuse and delay commitment to the correct region. In such cases, progressive refinement ensures eventual coverage, but the horizon constraint can prevent success.

\section{Conclusion and Future Work}
\label{sec:conclusion}

This work addressed autonomous object search in indoor environments under partial observability and perceptual uncertainty. We proposed a hybrid framework that combines Bayesian inference with DRL to unify uncertainty-aware belief estimation and adaptive decision making. The method maintains a spatial Dirichlet belief map over object classes, updated online from calibrated detections and background evidence, and learns a DQN policy that selects navigation goals over a hierarchical clustering-based abstraction.

Experimental results in two realistic Habitat~3.0 indoor environments show that belief-driven strategies substantially improve success rate over uninformed exploration. Moreover, the proposed BBDPS method achieves the highest success rates overall and yields the strongest efficiency gains in the larger environment, reducing search effort compared to a purely probabilistic utility-maximization baseline while preserving reliability. These results support the central conclusion of this work: explicitly modeling uncertainty and learning how to act on it can improve long-horizon search performance beyond either handcrafted heuristics or belief estimation alone.

Despite these gains, several limitations remain. The approach assumes a known occupancy grid and does not address joint mapping and semantic discovery, while performance is constrained by object detector quality, as false positives or missed detections can slow belief convergence and reduce exploration efficiency. In addition, the grid-based state representation requires training a separate policy per environment, limiting cross-scene generalization without further adaptation. Future work will focus on improving transfer across environments through more expressive state representations, incorporating semantic priors to guide belief refinement, and extending the framework to unknown or dynamic maps. We also plan to integrate active viewpoint selection with richer perception models and evaluate sim-to-real deployment.

\bibliographystyle{IEEEtran}
\bibliography{bibliography}

\end{document}